\documentclass[runningheads]{llncs}

\usepackage{graphicx}
\usepackage{amsfonts}

\begin{document}

\title{Learning-Augmented K-Means Clustering Using Dimensional Reduction}

\author{Issam K.O jabari\inst{1},
Shofiyah\inst{2}, 
Pradiptya Kahvi S\inst{3},
Novi Nur Putriwijaya\inst{4}, 
\and
Novanto Yudistira\inst{5}} 
\authorrunning{Jabari K.O. et al.}
\institute{Faculty of Computer Science, University of Brawijaya\inst{1}\inst{2}\inst{3}\inst{4}\inst{5}
\newline
\email{issam.jabari.2021@gmail.com}\inst{1}, \email{shofiyahpeace@student.ub.ac.id}\inst{2}, \email{pradiptya\_kahvi@student.ub.ac.id}\inst{3},
\email{nvnptrw@student.ub.ac.id}\inst{4},\newline \email{yudistira@ub.ac.id}\inst{5}}

\maketitle
\begin{abstract}
Learning augmented is a machine learning concept built to improve the performance of a method or model, such as enhancing its ability to predict and generalize data or features, or testing the reliability of the method by introducing noise and other factors. On the other hand, clustering is a fundamental aspect of data analysis and has long been used to understand the structure of large datasets. Despite its long history, the k-means algorithm still faces challenges.

One approach, as suggested by Ergun et al. \cite{b8}, is to use a predictor to minimize the sum of squared distances between each data point and a specified centroid. However, it is known that the computational cost of this algorithm increases with the value of k, and it often gets stuck in local minima. In response to these challenges, we propose a solution to reduce the dimensionality of the dataset using Principal Component Analysis (PCA).

It is worth noting that when using k values of 10 and 25, the proposed algorithm yields lower cost results compared to running it without PCA.

\textit{"Principal component analysis (PCA) is the problem of fitting a low-dimensional affine subspace to a set of data points in a high-dimensional space. PCA is well-established in the literature and has become one of the most useful tools for data modeling, compression, and visualization."}\cite{b18}
\end{abstract}

\keywords{K-Means, Principal Component Analysis, PCA, Dimensional Reduction, Predictor}

\section{Introduction}
Learning augmented is a machine learning concept designed to enhance the performance of a method or model, such as improving its ability to predict, generalize data, or handle noise. Augmented Learning, as a training model, enables the built model to learn data more reliably under various scenarios, even in the presence of unexpected noise \cite{b10}. These augmentations can take various forms, including applying transformations or adding noise to the input data.

Clustering is a popular data analysis technique used to group objects or data into clusters based on their characteristics \cite{b9}. Principal Component Analysis (PCA) and the k-means algorithm are commonly used methods for processing high-dimensional data. PCA can be applied as a technique to reduce dimensions before using the k-means algorithm to overcome high dimensionality challenges and improve data performance and interpretation \cite{b2}.

PCA transforms correlated variables into a new set of uncorrelated variables known as principal components \cite{b17}. Its goal is to preserve the maximum variance in the data by retaining the first principal components, which contain the most information \cite{b4}. PCA also helps transform data into a lower-dimensional space, reducing data complexity and extracting essential features \cite{b13}.

When using PCA on high-dimensional data, the first principal components are selected, containing the most variance in the original data. The data is then transformed into these new principal components using linear techniques such as Singular Value Decomposition or Eigenvalue Decomposition. Afterward, the k-means algorithm can be applied to the extracted principal components to cluster the data into different subsets based on similar patterns.

The k-means algorithm divides data into a predetermined number of categories or subgroups, denoted as "k" \cite{b11}. It relies on centroids or representative points for each category and aims to partition the data so that each point belongs to the cluster closest to its corresponding centroid. This process repeats until stability is achieved, and centroid changes stop.

Compared to k-means, PCA is effective in high-dimensional data cases due to its dimension-reduction capability and ability to handle common data variations more effectively. When PCA is used before applying k-means, it improves k-means' performance, reduces the impact of high dimensionality, and enhances data interpretation by identifying influential principal components. 

Ergun et al. \cite{b8} conducted research to address the problem of k-means clustering based on previously predicted labels. Predicted labels can be obtained from various clustering algorithms or supervised models, with added noise. They propose a polynomial-time procedure: for each matching component of the predicted labels, a strong average is computed (in a coordinate-wise manner). This strong average is the final clustering solution. The required time complexity is O(knd log n). 

This paper establishes theoretical guarantees on the approximation ratio of the solution, assuming limited label errors in predictor labels. To further improve running time, the authors use a dimension reduction technique of O(k/alpha) to cluster points in O(log n) dimensions and then obtain labels for the original data points using an approximate nearest neighbor data structure. This modified approach (Algorithm 3) has a running time of O(nd log n + poly(k, log n)) and achieves similar theoretical guarantees. Empirical evaluation on synthetic data and real datasets shows that the proposed method with k-means++ initialization outperforms classical k-means++ and remains competitive and robust even when predictor labels are corrupted.

After studying this research, we became aware of an ongoing problem when using k-means with the proposed predictor. The issue encountered is that as the value of k increases, even with k-means and additional algorithms as predictors, the performance tends to converge to local minima. This implies that increasing the value of k does not necessarily lead to optimal results. From this problem, we have learned that one possible solution to address local minima is to employ PCA to reduce the dimensionality of the dataset before applying clustering with the K-Means method.

\section{Related Works}
K-Means is one of the most commonly used algorithms in data clustering analysis, aimed at partitioning data into one or more clusters \cite{b19}. The K-Means algorithm operates by segregating data points based on their similarity in characteristics. Data points with similar characteristics are grouped into a single cluster, while those with dissimilar characteristics are allocated to separate clusters. This method offers the advantages of high efficiency, scalability, and rapid convergence, especially when dealing with substantial datasets. However, it has certain limitations, including the requirement to specify the number of clusters (K), the initial selection of cluster centers, and susceptibility to the influence of noise.

Previous research has utilized Principal Component Analysis (PCA) as a dimensionality reduction technique to enhance algorithmic performance. For instance, in a study by \cite{b20}, PCA was employed to reduce dataset dimensionality, followed by the application of Random Forest for intrusion detection in magma. Another study applied PCA for dimensionality reduction in the context of breast disease detection using Artificial Neural Network (ANN) \cite{b16}, as well as for diabetes detection using Linear Discriminant Analysis (LDA) \cite{b6}.

Based on the findings of previous research, which compared datasets using PCA with those without PCA, it is evident that utilizing PCA for dimensionality reduction significantly enhances accuracy when compared to datasets without PCA-based dimensionality reduction.

\section{Methods}

\subsection{Learning-Augmented}
Learning-Augmented Clustering refers to a methodology that incorporates machine learning techniques to enhance the clustering process. Clustering is the task of grouping similar data points together based on their inherent patterns or similarities. Traditional clustering algorithms, such as k-means or hierarchical clustering, often rely on predefined distance metrics or similarity measures to assign data points to clusters.

In Learning-Augmented Clustering, machine learning techniques are utilized to improve the clustering process by learning the underlying patterns and relationships within the data \cite{b21}. This approach typically involves combining traditional clustering algorithms with machine learning models, such as neural networks or support vector machines, to enhance the accuracy and effectiveness of clustering.

The learning component in Learning-Augmented Clustering can encompass various aspects, including feature selection, dimensionality reduction, or even training a separate machine learning model to assist in the clustering process. These techniques aim to leverage the power of machine learning to identify complex patterns, handle high-dimensional data, and manage noisy or incomplete data, thereby enhancing overall clustering performance.

By incorporating learning into the clustering process, Learning-Augmented Clustering algorithms can adapt to diverse data types and capture more intricate relationships among data points. This approach has applications in various domains, such as image and text analysis, customer segmentation, anomaly detection, and bioinformatics, where traditional clustering algorithms may struggle to effectively handle complex data structures.

In this section, we conduct a performance evaluation of the K-means algorithm in conjunction with a Predictor. We compare the algorithm's performance before and after applying dimensionality reduction using PCA. To calculate the cost of error, we utilize various datasets and employ different predictors. Through the assessment of the algorithm's performance, we aim to comprehend the impact of incorporating a Predictor within the K-means framework and investigate the effectiveness of dimensionality reduction using PCA in improving the algorithm's accuracy.

To conduct our evaluation, we employ diverse datasets that cover a range of scenarios and complexities, enabling us to assess the algorithm's robustness and generalizability across different domains. Additionally, we employ various predictors in conjunction with the K-means algorithm, allowing us to explore the influence of different modeling techniques on the algorithm's overall performance and identify the most suitable predictor for a given dataset and task.

Through this comprehensive evaluation, we aim to gain insights into the strengths and limitations of the K-means algorithm when combined with a Predictor and understand how dimensionality reduction using PCA can contribute to improving the algorithm's accuracy and reducing the cost of error.

\subsection{Preliminaries:} We use $[n]$ to denote the set $\{1, 2, \ldots, n\}$. Given the set of cluster centers $C$, we can partition the input points $P$ into $k$ clusters $C_1, C_2, \ldots, C_k$ according to the closest center to each point. If a point is grouped in $C_i$ in the clustering, we refer to its label as $i$. Note that labels can be arbitrarily permuted as long as the labeling across the points of each cluster is consistent. 

It is well-known that in k-means clustering, the $i$-th center is given by the coordinate-wise mean of the points in $C_i$. Given $x \in \mathbb{R}^d$ and a set $C \subset \mathbb{R}^d$, we define $d(x; C) = \min_{c\in C} \|x - c\|$. Note that there may be many approximately optimal clusterings, but we consider a fixed one for our analysis.

\subsection{Datasets}
The following three datasets were used:
\begin{enumerate}
\item \textbf{The Oregon Graph}: dataset consists of nine graphs that represent the peering information between Autonomous Systems (AS) as inferred from Oregon route-views. These graphs were captured over a period between March 31, 2001, and May 26, 2001. Each graph snapshot provides a different view of the AS peering relationships during that time frame. The number of nodes in each graph ranges from a minimum of 10,670 to a maximum of 11,174, indicating the varying complexity and size of the AS networks captured in the dataset \cite{b3}.
\item \textbf{The PHY dataset :}is sourced from the KDD Cup 2004 \cite{b1}, a prestigious data mining competition. For our specific study, we extract a dataset comprising 104 random samples.
This file contains the training data for the quantum physics task. It consists of 50,000 cases specifically designed for training purposes in the quantum physics domain  .
\item \textbf{CIFAR-10 dataset:} The CIFAR-10 dataset comprises 6000 examples for each of the 10 classes, resulting in a total of 60,000 labeled examples \cite{b14}.
\end{enumerate}
\subsection{Algorithm}
It is shown in Table \ref{tab:predictorclustering}  Algorithm of Predictor Clustering
\begin{table}[h!]
\extracolsep{\fill}
\def\arraystretch{1.25}%
\centering
\caption{Algorithm of Predictor Clustering}
\label{tab:predictorclustering}
\resizebox{\columnwidth}{!}{
    \begin{tabular}[t]{@{}l@{}}
    \hline
    \textbf{ALGORITHM}: PredictorClustering \\
    \hline
    \textbf{TRAINING\_PREDICTOR()} // Train or obtain the predictor \\
    \hline
    \textbf{INPUT\_DATA} $\leftarrow$ \textbf{GET\_INPUT\_DATA()} // Get the input data \\
    \hline
    \textbf{REDUCED\_DATA} $\leftarrow$ \textbf{PCA\_REDUCTION}(INPUT\_DATA) // Reduce the input dataset using PCA
    \textbf{PREDICTIONS} $\leftarrow$ [] \\
    \textbf{FOR EACH} data\_point \textbf{IN} INPUT\_DATA \textbf{DO} \\
    \quad prediction $\leftarrow$ \textbf{PREDICT}(data\_point) // Use the predictor to make predictions \\
    \quad \textbf{PREDICTIONS.APPEND}(prediction) // Store the predictions \\
    \textbf{LABEL\_ASSIGNMENT}(INPUT\_DATA, PREDICTIONS) // Assign the predictions to the data points \\
    \textbf{CLUSTER\_ASSIGNMENTS} $\leftarrow$ PREDICTIONS // Use the predictions as initial cluster assignments \\
    \textbf{REPEAT} \\
    \quad \textbf{UPDATE\_CLUSTER\_CENTERS}(CLUSTER\_ASSIGNMENTS) // Update the cluster centers based on current assignments \\
    \quad \textbf{NEW\_CLUSTER\_ASSIGNMENTS} $\leftarrow$ [] \\
    \quad \textbf{FOR EACH} data\_point \textbf{IN} INPUT\_DATA \textbf{DO} \\
    \quad \quad nearest\_cluster $\leftarrow$ \textbf{FIND\_NEAREST\_CLUSTER}(data\_point, CLUSTER\_ASSIGNMENTS) // Find the nearest cluster center \\
    \quad \quad \textbf{NEW\_CLUSTER\_ASSIGNMENTS.APPEND}(nearest\_cluster) // Assign the data point to the nearest cluster \\
    \quad \textbf{END FOR} \\
    \quad \textbf{IF} \textbf{NEW\_CLUSTER\_ASSIGNMENTS} == \textbf{CLUSTER\_ASSIGNMENTS} \textbf{THEN} \\
    \quad \quad \textbf{EXIT REPEAT} // Convergence reached, exit the loop \\
    \quad \textbf{END IF} \\
    \quad \textbf{CLUSTER\_ASSIGNMENTS} $\leftarrow$ \textbf{NEW\_CLUSTER\_ASSIGNMENTS} // Update the cluster assignments \\
    \textbf{UNTIL CONVERGENCE} \\
    \textbf{FINAL\_CLUSTERS} $\leftarrow$ \textbf{GENERATE\_CLUSTERS}(INPUT\_DATA, CLUSTER\_ASSIGNMENTS) // Generate the final clusters \\
    \textbf{OUTPUT\_RESULTS}(FINAL\_CLUSTERS) // Output the final clustering result \\
    \textbf{K\_MEANS\_CLASSIFICATION}(FINAL\_CLUSTERS) // Perform classification using K-Means algorithm \\
    \hline
    \end{tabular}
}
\end{table}

\subsection{Predictors}
\textbf{Predictors} can be used to help group data in datasets that have additional information. In addition, predictors can also be made from a class of datasets that vary over time, such as census data or spectral clustering for temporal graphs. Predictors can experience adversarial errors which can affect the quality of clustering.

The authors \cite{b8} was modified k-means clustering algorithm with predictors is used to find $k$ centers that minimize certain objectives and assign each point to one of these $k$ centers. The predictor used is a neural network that has been trained on the CIFAR-10 dataset. This predictor is used to label each data point in the dataset that will be clustered using the k-means algorithm.

In our experiments, we employ the following predictors:
\begin{enumerate}
\item \textbf{The Nearest Neighbor predictor:} For the Oregon dataset, we utilize the Nearest Neighbor predictor. This predictor is specifically employed to determine the best clustering of the node embeddings in Graph 1. To achieve this, we apply Lloyd's algorithm, running multiple iterations until convergence, starting with an initial seeding process using the k-means++ algorithm \cite{b8}.\\
The Nearest Neighbor predictor operates by taking as input a point in R2, which represents a node embedding from any of the subsequent eight graphs. It then outputs the label of the node in Graph 1 that is closest to the given point. This proximity-based approach enables us to assign labels to the nodes in the later graphs based on their proximity to the nodes in Graph 1.
\item \textbf{ The Noisy predictor:} Serves as the primary predictor for the PHY dataset. To construct this predictor, we begin by obtaining the best solution for k-means clustering on our datasets. This involves an initial seeding step using the k-means++ algorithm, followed by multiple iterations of Lloyd's algorithm until convergence is achieved.\\
Once we have obtained the initial clustering solution, we introduce noise into the predictor by randomly corrupting the resulting labels. Each label is independently changed to a uniformly random label, with the error probability varying from 0 to 1. \\
In our analysis, we compare the performance of clustering using these noisy labels alone against the approach of processing these labels using the k-means algorithm. The cost of clustering is then reported for both scenarios, allowing us to evaluate the impact of noisy labels on the effectiveness of the k-means algorithm.
\item \textbf{The Neural Network predictor:} Employed in our experiments is based on a standard architecture known as ResNet18. This neural network model has been trained on the training portion of the CIFAR-10 dataset and serves as the oracle for the testing portion used in our experiments. \\
To ensure consistency and reproducibility, we utilize a pre-trained ResNet18 model obtained from Huy Phan \cite{b15}. It's worth noting that the primary task of the neural network is to predict the class of the input image. However, it's important to acknowledge that the assigned class value is strongly correlated with the optimal k-means cluster group. This correlation indicates that the class predictions made by the neural network can provide valuable insights into the clustering behavior and effectiveness of the k-means algorithm.

\end{enumerate}

\subsection{Principal Component Analysis (PCA)}
Dimensionality reduction algorithms are techniques used to reduce the number of features or variables in a dataset while preserving important information. These algorithms help overcome the curse of dimensionality, where high-dimensional data can lead to increased computational complexity, overfitting, and difficulties in visualization and interpretation. One way of dimensionality reduction is through feature extraction. Feature extraction methods transform the original high-dimensional data into a lower-dimensional representation by creating new features (or components) that capture most of the information from the original features. These methods aim to maintain relevant information while reducing the dimensionality. Among the existing feature extraction methods, Principal Component Analysis (PCA) is a well-known technique.
Principal Component Analysis (PCA), also known as the Karhunen-Loeve method or K-L, seeks orthogonal k-dimensional vectors that best represent the data, where $k \leq n$. In short, PCA linearly transforms the data into a set of new uncorrelated variables called principal components, which are then ordered by their importance. The original data is then projected onto this much smaller space, resulting in dimensionality reduction \cite{b5}.

PCA "combines" attribute essence by creating alternative, smaller sets of variables \cite{b12}. The original data can then be projected onto this reduced set. PCA often reveals previously unsuspected relationships and enables interpretations that are typically not obtained.

The basic procedure is as follows \cite{b5}:

\begin{enumerate}
\item The input data is normalized, so that each attribute is within the same range. This step helps ensure that attributes with larger domains do not dominate those with smaller domains.
\item PCA computes k orthonormal vectors that provide a basis for the normalized input data. These are unit vectors that are mutually perpendicular in direction. These vectors are called principal components, and the input data is a linear combination of the principal components.
\item The principal components are sorted in decreasing order of "significance" or strength. The principal components essentially serve as a new set of axes for the data, providing important information about variance. In other words, the axes are ordered in such a way that the first axis captures the most variance among the data, the second axis captures the next highest variance, and so on.
\item Since the components are sorted in decreasing "significance," the size of the data can be reduced by eliminating weaker components, i.e., components with low variance. By using the strongest principal components, it should be possible to reconstruct a good approximation of the original data.
\end{enumerate} PCA can be applied to both ordered and unordered attributes and can handle sparse and skewed data. Multidimensional data with more than two dimensions can be handled by reducing the problem to two dimensions. The principal components can be used as input for multiple regression and cluster analysis. Compared to wavelet transformation, PCA tends to perform better with sparse data, while wavelet transformation is more suitable for high-dimensional data \cite{b7}.
Equations involved:
\begin{enumerate}
\item Calculate the mean vector of the matrix \(X\):

\begin{eqnarray}
\bar{X} = \frac{1}{m} \sum_{i=1}^{m} X_i \}
\end{eqnarray}

\item Apply the transformation \(X - \bar{X}\):

\begin{eqnarray}
X'= X - \bar{X} \end{eqnarray}

\item Calculate the scatter matrix C:

\begin{eqnarray}
C = \frac{1}{m} X'^T X' 
\end{eqnarray}

\item Compute the eigenvalues and eigenvectors:
We use eigende composition to compute them.

First, calculate the eigenvector matrix \(V\) consisting of the principal component vectors:
\begin{eqnarray}
V = eigenvectors(C) 
\end{eqnarray}

Then, compute the eigenvalue matrix \(D\) present in the diagonalized form:
\begin{eqnarray}   
D = eigenvalues(C) 
\end{eqnarray}

\item Sort the eigenvectors and eigenvalues:
The principal component vectors and eigenvalues are sorted in descending order according to the eigenvalues.

\item Select the principal components:
We can choose a certain number of principal components to retain based on the desired explained variance ratio. In this example, we will retain the top two principal components only.

\item Calculate the explained variance ratio:
The explained variance ratio is the ratio of each principal component's eigenvalue to the sum of all eigenvalues.

\begin{eqnarray}  
Explained Variance Ratio = \frac{D_k}{\sum_{i=1}^{k} D_i} 
\end{eqnarray}
\end{enumerate}
The new matrix is smaller in dimensions, but it carries the same characteristics of the original matrix, which makes it easier to deal with it in any other algorithm

\subsection{K-means}
\textbf{K-Means} is a popular clustering algorithm used to partition data into one or more clusters \cite{b19}. The algorithm works by iteratively assigning data points to clusters based on their similarity in characteristics and updating the cluster centers.

The steps of the K-Means algorithm can be summarized as follows:

\begin{enumerate}
    \item Initialize the number of clusters $K$ and randomly assign $K$ initial cluster centers.
    \item Assign each data point to the cluster whose center is closest to it based on a distance metric (usually Euclidean distance).
    \item Recalculate the cluster centers by taking the mean of all data points assigned to each cluster.
    \item Repeat steps 2 and 3 until convergence criteria are met, such as the maximum number of iterations or when the cluster assignments no longer change significantly.
\end{enumerate}
The final result is a set of $K$ clusters, where each cluster contains data points with similar characteristics.

Algorithm Equations:
\begin{enumerate}
\item Randomly initialize the cluster centers:
\begin{eqnarray} 
C^0 = Rand(K)
\end{eqnarray}

\item Compute the distances between the data points and the cluster centers:
\begin{eqnarray} 
D = Dist(X, C^t)
\end{eqnarray}

\item Assign the data points to the nearest cluster center:
\begin{eqnarray} 
A = Assign(D)
\end{eqnarray}

\item Update the cluster centers:
\begin{eqnarray} 
C^{(t+1)} = Update(X, A, K)
\end{eqnarray}

\item Check for algorithm convergence:
\begin{eqnarray} 
Conv = CheckConv(C^t, C^{t+1})
\end{eqnarray}

\item Repeat the algorithm until convergence:
while not Conv from point 2 to 5.

\item Output the final assignments and cluster centers:
\begin{eqnarray} 
output(A, C^{(t+1)})
\end{eqnarray}
\end{enumerate}

K-Means has several advantages, including its efficiency and scalability, making it suitable for large datasets. It also has fast convergence when processing a large amount of data. However, the algorithm has some limitations. It requires the initialization of the number of clusters $K$, which can be challenging to determine in advance. Additionally, K-Means relies on the random selection of initial cluster centers, which may lead to different results with each run. The algorithm is also sensitive to the influence of outliers and noise in the data.

Overall, K-Means is a widely used algorithm in clustering analysis due to its simplicity and effectiveness. It is applicable in various domains such as data mining, pattern recognition, and customer segmentation, among others.



\section{Result}
In this stage, the initial step is to preprocess the data by reducing its dimension using PCA. The dimensionality of the data can be observed in Figures\ref{fig:Dataset_Dimension1} and \ref{fig:Dataset_Dimension2} below:

\begin{figure}[htp]
    \centering
    \includegraphics[width=6cm]{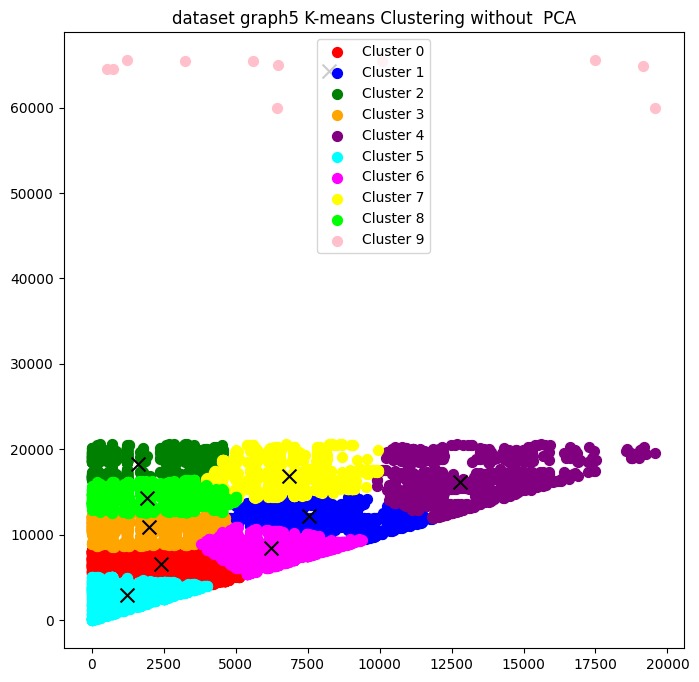}
    \caption{Plot of Oregon Dataset Dimension without Using PCA}
    \label{fig:Dataset_Dimension1}
\end{figure}

\begin{figure}[htp]
    \centering
    \includegraphics[width=6cm]{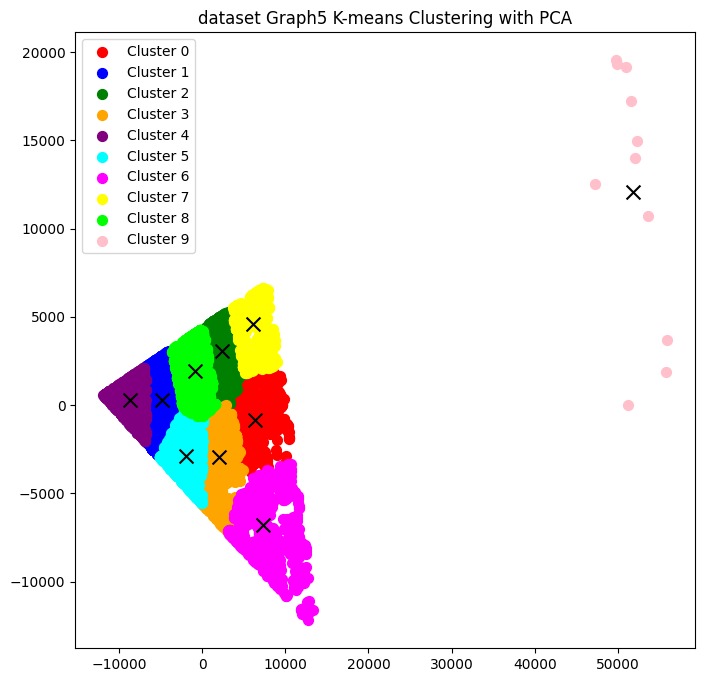}
    \caption{Plot of Oregon Dataset Dimension Using PCA}
    \label{fig:Dataset_Dimension2}
\end{figure}

Based on the above figures, it is evident that the dimension of the data has been significantly reduced. Initially, the data ranged from above 0 up to 60,000. After applying PCA, the dimension of the data decreased, with values ranging from around -10,000 to a maximum of 20,000. The "X" in the figure indicates the midpoint (cluster) in each class.

\begin{figure}[htp]
    \centering
    \includegraphics[width=6cm]{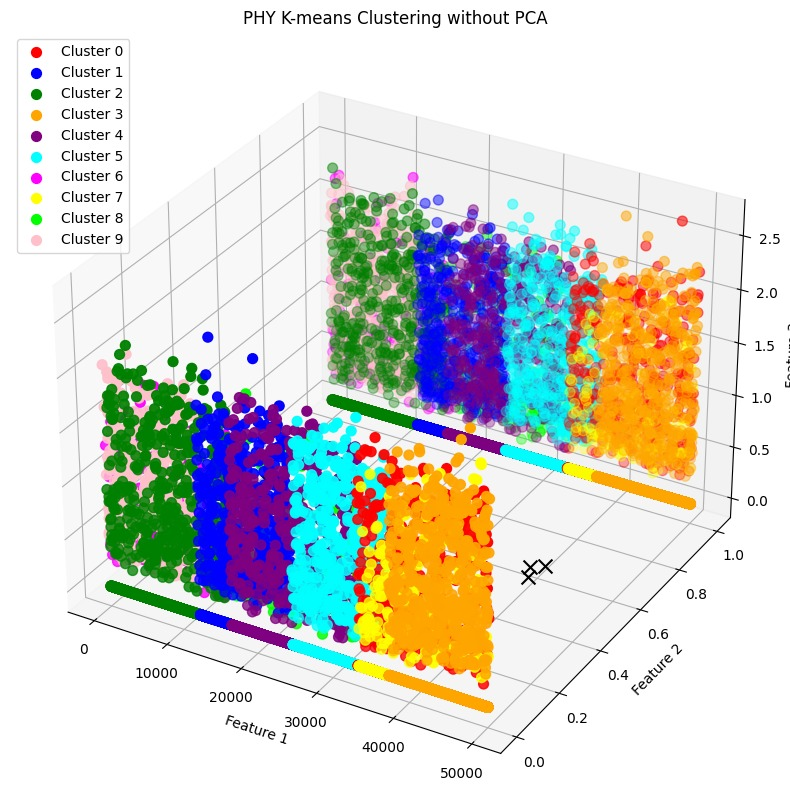}
    \caption{Plot of PHY Dataset Dimension without Using PCA}
    \label{fig:Dataset_Dimension3}
\end{figure}

\begin{figure}[htp]
    \centering
    \includegraphics[width=6cm]{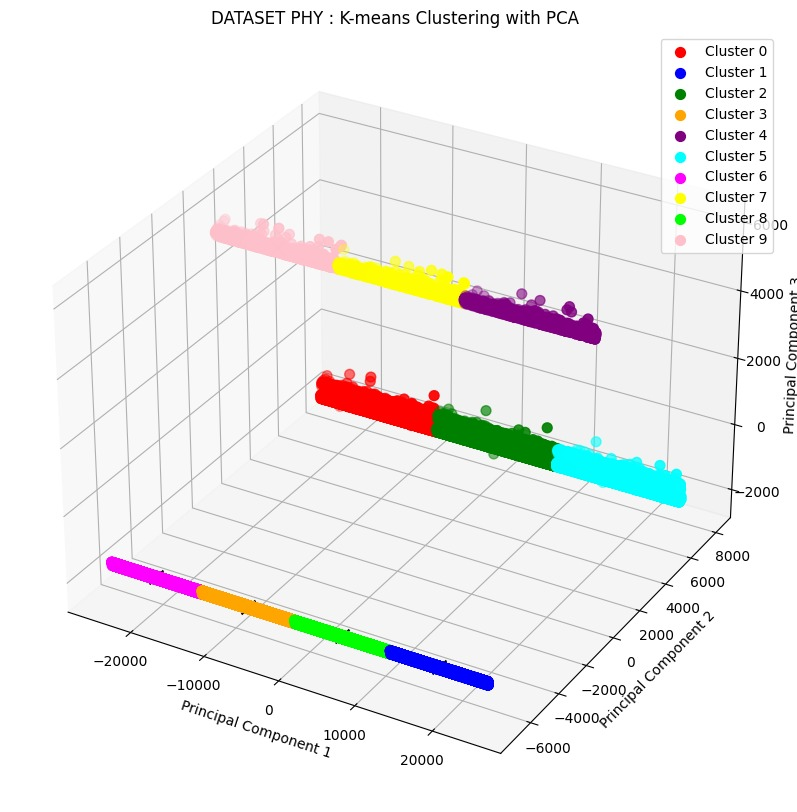}
    \caption{Plot of PHY Dataset Dimension Using PCA}
    \label{fig:Dataset_Dimension4}
\end{figure}

\begin{figure}[htp]
    \centering
    \includegraphics[width=6cm]{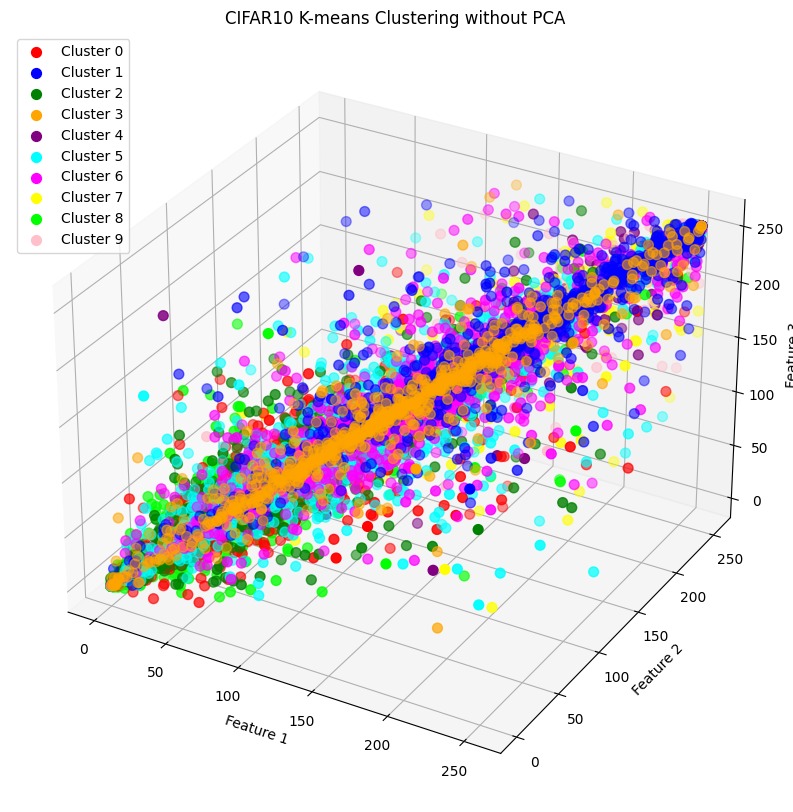}
    \caption{Plot of CIFAR10 Dataset Dimension without Using PCA}
    \label{fig:Dataset_Dimension5}
\end{figure}

\begin{figure}[htp]
    \centering
    \includegraphics[width=6cm]{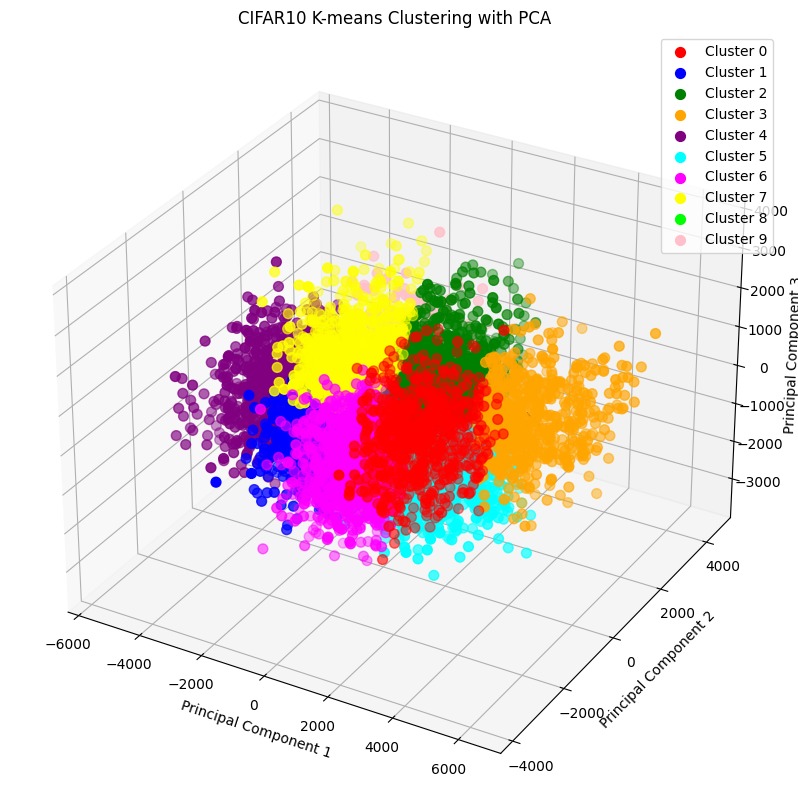}
    \caption{Plot of CIFAR10 Dataset Dimension Using PCA}
    \label{fig:Dataset_Dimension6}
\end{figure}

After dimensionality reduction using PCA, the dataset was applied to the k-means algorithm and the proposed Algorithm 3 by Ergun et al. Figures \ref{fig:Dataset_Dimension7} and \ref{fig:Dataset_Dimension8} below show the comparison of the performance between the initial k-means and the k-means with PCA:

\begin{figure}[htp]
    \centering
    \includegraphics[width=6cm]{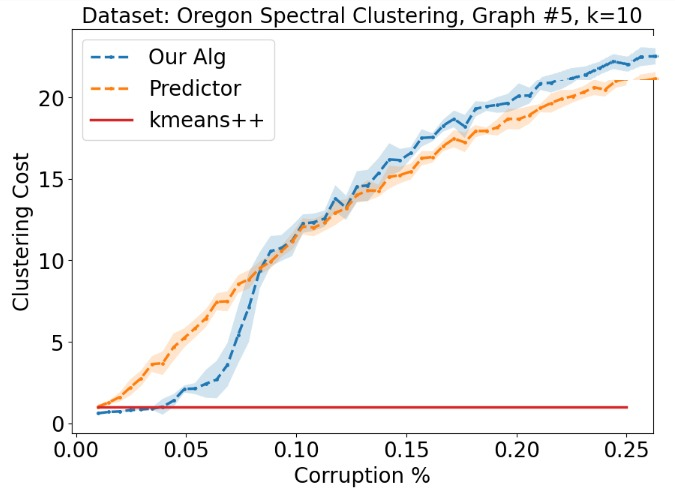}
    \caption{Algorithm Cost Graph on Oregon Dataset without PCA with k=10}
    \label{fig:Dataset_Dimension7}
\end{figure}

\begin{figure}[htp]
    \centering
    \includegraphics[width=6cm]{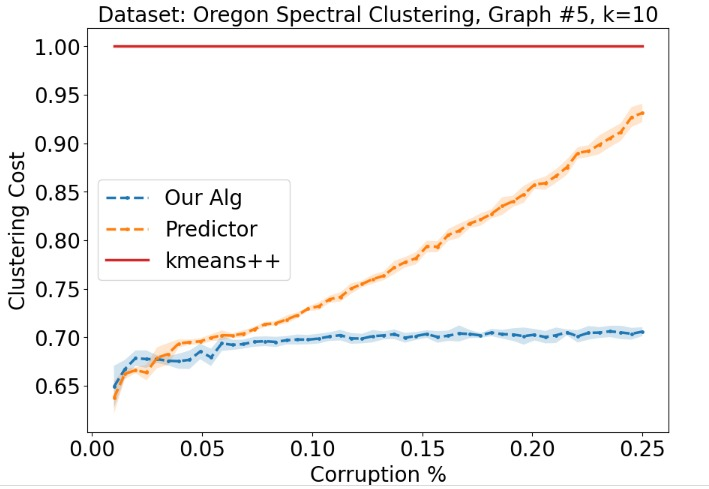}
    \caption{Algorithm Cost Graph on Oregon Dataset Using PCA with k=10}
    \label{fig:Dataset_Dimension8}
\end{figure}

Figure \ref{fig:Dataset_Dimension7} represents the algorithm cost generated from the use of k-means and the proposed predictor. It is observed that the resulting algorithm cost is always above 1 and continues to increase to more than 20 as the computational time depicted by the blue line on the graph. In comparison, Figure \ref{fig:Dataset_Dimension8} shows that the resulting algorithm cost is always below 1 and remains stable even as the computational time increases, depicted by the blue line on the graph.\\
Taking this into account, clustering cost calculations were also performed with a value of k=25 to further investigate whether the increased value of k would result in better performance for k-means compared to local k-means. The clustering cost calculation with k=25 is shown below:

\begin{figure}[htp]
    \centering
    \includegraphics[width=6cm]{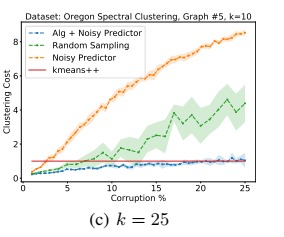}
    \caption{Algorithm Cost Graph on Oregon Dataset Using PCA with k=25}
    \label{fig:Dataset_Dimension9}
\end{figure}

From the graph in Figure \ref{fig:Dataset_Dimension9}, it is observed that with an increased value of k, namely 25, the clustering cost starts below 1, then increases to 1.1. However, as corruption increases, the performance and clustering cost remain stable without significant increases. In other words, compared to previous studies, the clustering cost significantly decreases, and the clustering cost value is lower than that of classical k-means.

\begin{figure}[htp]
    \centering
    \includegraphics[width=6cm]{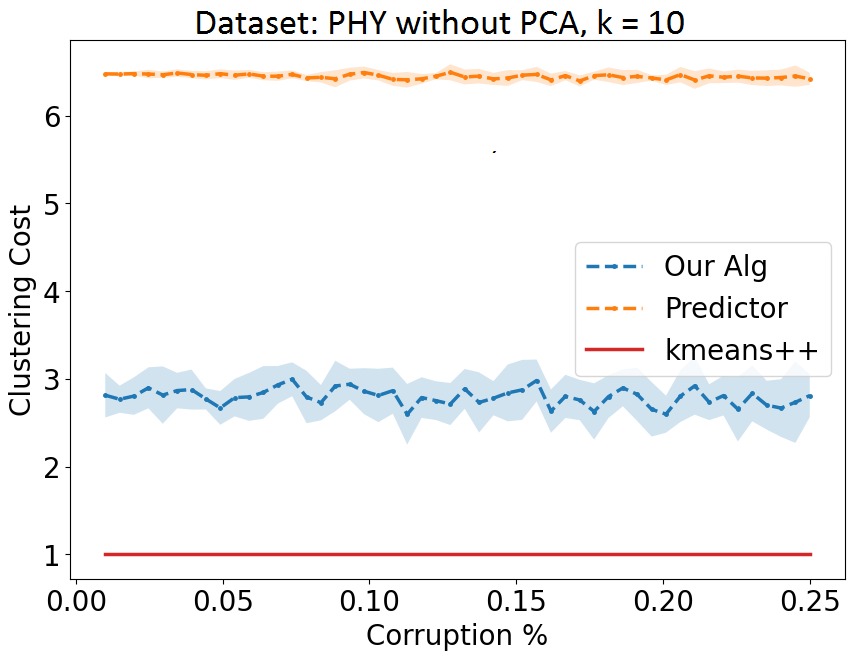}
    \caption{Algorithm Cost Graph on PHY Dataset without PCA with k=10}
    \label{fig:Dataset_Dimension10}
\end{figure}

\begin{figure}[htp]
    \centering
    \includegraphics[width=6cm]{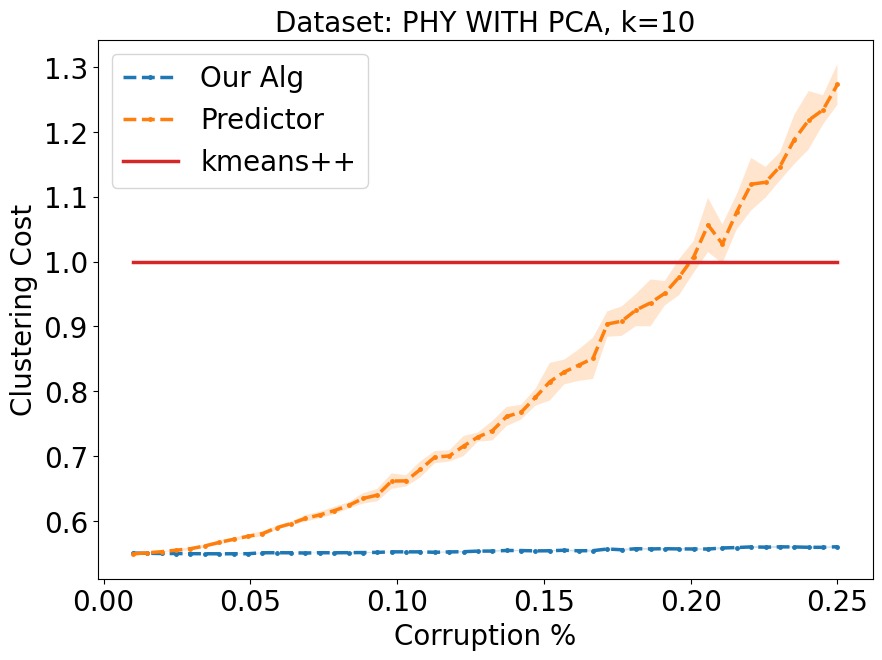}
    \caption{Algorithm Cost Graph on PHY Dataset Using PCA with k=10}
    \label{fig:Dataset_Dimension11}
\end{figure}

Based on the figure \ref{fig:Dataset_Dimension10} and \ref{fig:Dataset_Dimension11} above, it is known that after reducing the data with PCA, the performance of k-means and predictor becomes more efficient based on the clustering cost spent.

\begin{figure}[htp]
    \centering
    \includegraphics[width=6cm]{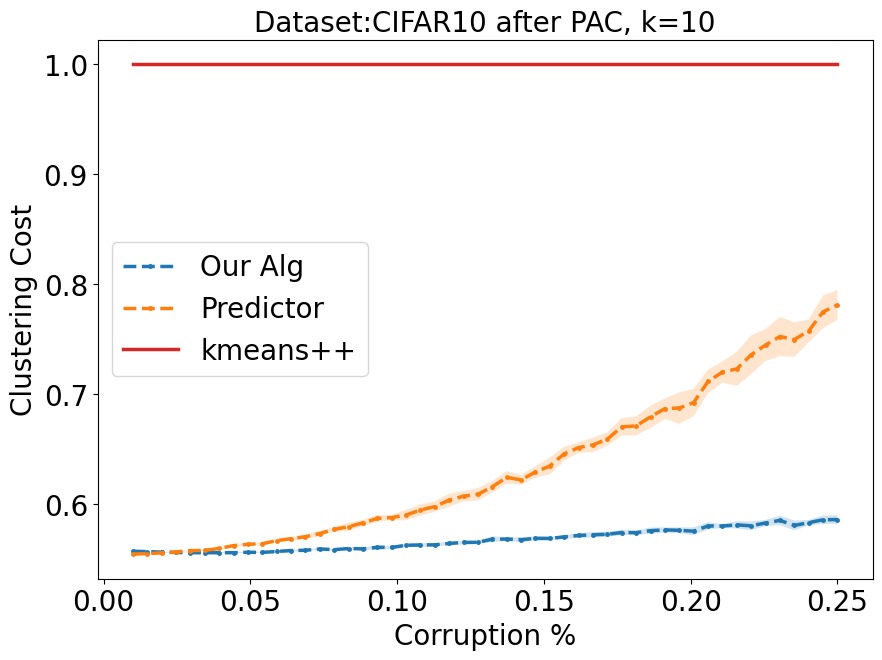}
    \caption{Algorithm Cost Graph on CIFAR10 Dataset without PCA with k=10}
    \label{fig:Dataset_Dimension12}
\end{figure}

In line with the CIFAR10 dataset \ref{fig:Dataset_Dimension12}, the clustering cost is not as high as before using PCA. In other words, reducing the dataset before predictor and k-means can improve efficiency and eliminate the local minima problem due to the increasing value and always above 1.


\section{Conclusion}

In conclusion, our analysis and findings underscore the significance of using the k-means algorithm for data clustering and facilitating data comprehension. K-means is particularly suitable when dealing with low-dimensional data and clear data structures that can be clustered effectively.

However, k-means can become less efficient when dealing with high-dimensional data, where handling these high dimensions and understanding complex data structures can become extremely challenging. In such cases, Principal Component Analysis (PCA) can be used to reduce dimensionality and enhance the performance of k-means. PCA analyzes the data and extracts the principal components that carry the most information, simplifying the data and making it more amenable to clustering using k-means.

It's worth noting that PCA may not be effective when dealing with low-dimensional or less complex data, as its application in such cases may lead to the loss of valuable information. Therefore, it is essential to evaluate the nature and complexity of the data before deciding to use PCA.

In summary, using k-means with PCA is a powerful option when dealing with high-dimensional and complex data, while PCA may not be beneficial when dealing with low-dimensional or straightforward data. Striking the right balance by employing suitable tools for the specific data type can enhance overall data clustering performance and provide deeper insights into the information.


\bibliographystyle{unsrt}

\end{document}